\documentclass[letterpaper, 10 pt, conference]{ieeeconf}  

\IEEEoverridecommandlockouts                              

\usepackage{graphicx}
\usepackage{amsmath} 
\usepackage{subfig}
\usepackage{algpseudocode}
\usepackage{algorithm}
\usepackage{booktabs}
\usepackage{xspace}
\usepackage{multirow}
\usepackage{dblfloatfix} 
\usepackage{url}            
\usepackage{hyperref}
\usepackage{amsfonts}
\usepackage{xcolor}
\usepackage{todonotes}

\title{\LARGE \bf
HyperPPO: A scalable method for finding small policies for robotic control 
}

\author{Shashank Hegde, Zhehui Huang and Gaurav S. Sukhatme\\
University of Southern California
\thanks{\footnotesize{\tt khegde|zhehuihu|gaurav@usc.edu}}
\thanks{GSS holds concurrent appointments as a Professor at USC and as an Amazon Scholar. This paper describes work performed at USC and is not associated with Amazon.}%
}

\begin{document}

\maketitle

\thispagestyle{empty}
\pagestyle{empty}

\newcommand{\etc}{\emph{etc.}\xspace}
\newcommand{\ie}{\emph{i.e.,}\xspace}
\newcommand{\eg}{\emph{e.g.,}\xspace}
\newcommand{\etal}{\emph{et al.}\xspace}
\newcommand{\wrt}{with respect to\xspace}
\newcommand{\shs}[1]{\textcolor{red}{[Shashank says: #1]}}         

\begin{abstract}

Models with fewer parameters are necessary for the neural control of memory-limited, performant robots. 
Finding these smaller neural network architectures can be time-consuming.
We propose HyperPPO, an on-policy reinforcement learning algorithm that utilizes graph hypernetworks to estimate the weights of multiple neural architectures simultaneously.
Our method estimates weights for networks that are much smaller than those in common-use networks yet encode highly performant policies.
We obtain multiple trained policies at the same time while maintaining sample efficiency and provide the user the choice of picking a network architecture that satisfies their computational constraints.  
We show that our method scales well - more training resources produce faster convergence to higher-performing architectures.
We demonstrate that the neural policies estimated by HyperPPO are capable of decentralized control of a Crazyflie2.1 quadrotor.
Website: \href{https://sites.google.com/usc.edu/hyperppo}{https://sites.google.com/usc.edu/hyperppo}

\end{abstract}

\section{Introduction}
\label{sec:intro}

A common practice in robot learning (particularly deep reinforcement learning) is to fix a network size and architecture and train it to approximate the near-optimum policy for a given task. 
For locomotion tasks with only proprioceptive sensing, networks of $\sim 256$ neurons and $\sim  3$ layers are commonly employed~\cite{haarnoja2019learning}, while for exteroceptive sensing, the configuration of the network varies with the data modality~\cite{pmlr-v164-yu22a}. 
For tasks that require the neural network controller to be deployed onto a real robot, especially one with memory and computational constraints such as the Crazyflie2.1, with which we experiment here (192Kb of onboard RAM)~\cite{giernacki2017crazyflie}, the choice of network size and architecture is of paramount importance.

\begin{figure}
    \centering
    \includegraphics[width=\columnwidth]{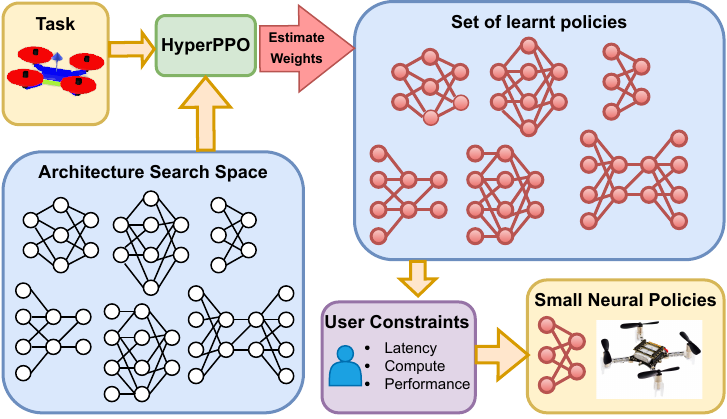}
    \caption{For a given task and a large architecture search space, HyperPPO learns to estimate weights for multiple architectures simultaneously. The user can choose an architecture based on their performance requirements and computational constraints from the set of learned policies.}
    \label{fig:framework}
    \vspace{-4mm}
\end{figure}

There has been significant recent progress in neural architecture search (NAS)~\cite{white2023neural}. However, this has not focused on applications to neural robotic control. The problem of finding small yet performant neural networks for robot control is further exacerbated by the fact that performance and size of neural networks are not directly correlated~\cite{hegde2023efficiently}. Here, we build on the approach in~\cite{hegde2023efficiently} and present a method (Figure \ref{fig:framework}) that trains thousands of architecturally unique neural control policies simultaneously. We give the user the ability to choose an architecture that fits within their computation constraints and meets their performance requirements. We note that post-training, the weights for any chosen architecture can be estimated in one forward pass of our trained model.
    
\noindent {\bf Contributions:} The method proposed in \cite{hegde2023efficiently} is off-policy. Such methods tend to be sample-efficient yet time-inefficient in training (when one measures wall-clock training time). Here we present an on-policy method (HyperPPO) that 
simultaneously produces thousands of policies, each with a unique architecture. HyperPPO has sample efficiency similar to one training run of regular proximal policy optimization (PPO) and results in unique performant policies for each architecture. We propose two versions of HyperPPO: with vectorized standard deviations (HyperPPO-VSD), suitable for the setting when training data are abundant and a fast simulator is available, and with common standard deviation (HyperPPO-CSD), suitable in the setting when gathering data is harder. We analyze and ablate the trade-offs of each version. We benchmark HyperPPO-VSD on GPU accelerated environments and HyperPPO-CSD on the quadrotor simulator, QuadSwarm~\cite{huang2023quadswarm}. We show that small networks estimated by HyperPPO-VSD are capable of outperforming the same networks obtained by training with regular PPO.
We also show that the weights estimated by HyperPPO-CSD for a tiny neural network (just one hidden layer with 4 neurons) can be successfully deployed on a Crazyflie2.1 for autonomous flight control.


\section{Related Work}
\label{sec:related_works}

\subsection{Proximal Policy Optimization (PPO)}
\label{sec:related_works_ppo}
    PPO is a widely adopted on-policy learning algorithm~\cite{schulman2017proximal}. As opposed to off-policy learning algorithms, PPO provides separate loops for sample collection and training. This separation allows for massive parallelization, which provides trained policies more quickly. Further, PPO has been shown to have better stability. The governing equations of PPO are as follows.
    
        \begin{center}
            $\begin{aligned}
            r_t(\theta) &= \frac{\pi_\theta\left(a_t \mid s_t\right)} {\pi_{\theta_k}\left(a_t \mid s_t\right)} \\
            \hat{A}_t^{\pi_\theta} &=\delta_t+(\gamma \lambda) \delta_{t+1}+\ldots+(\gamma \lambda)^{T-t+1} \delta_{T-1} \\
            \text { where } \delta_t &=r_t+\gamma V^{\pi_\theta}\left(s_{t+1}\right)-V^{\pi_\theta}\left(s_t\right) 
            \end{aligned}$\\
            
            $\mathcal{L}_{\theta_k}(\theta)=\underset{\tau \sim \pi_\theta}{\mathrm{E}}\left[\min \left(r_t(\theta) \hat{A}_t^{\pi_{\theta_k}}, \operatorname{clip}\left(r_t(\theta), 1\pm \epsilon\right) \hat{A}_t^{\pi_{\theta_k}}\right)\right]$
        \end{center}

    $r_t(\theta)$ is the importance sampling ratio function between the policy that is used to collect data and the $k$'th version of the policy. $V^{\pi_\theta}\left(s_t\right)$ is the value function estimated by the critic for the policy $\pi_\theta$ at the state $s_t$.The generalized advantage estimate is given by $\hat{A}_t^{\pi_\theta}$. Finally, $\mathcal{L}_{\theta_k}(\theta)$ is the clipped loss objective.

Off-policy methods tend to be slower than on-policy methods, as the latter can be optimized easily. 
Further, on-policy methods have fewer hyperparameters and can have higher convergence stability if we have sufficient environment instances~\cite{andrychowicz2021matters}. Optimizations needed to improve the performance of PPO are documented in~\cite{shengyi2022the37implementation}. A benefit of using PPO is the ability to scale with more computational resources. The availability of highly parallelized environments~\cite{weng2022envpool} and GPU-based physics engines~\cite{freeman2021brax,makoviychuk2021isaac}, have been shown to work well with PPO~\cite{rudin2021learning,weng2022envpool}. For exploration, PPO generally samples its actions from a stochastic policy. The mean is obtained as the output from a parameterized state-conditioned network. The standard deviation is obtained either with another state-conditioned network or is simply characterized as a (non-state-conditioned) array whose values are directly modified during training. Here, we will consider the later version.

\subsection{Neural Architecture Search}
\label{sec:related_works_nas}
Neural architecture search~\cite{white2023neural} is the process of searching for an optimal neural architecture for a given task.
While reinforcement learning has been used for NAS \cite{zoph2016neural}, the use of NAS for reinforcement learning-based policies is still an under-explored area.
NAS has tremendous opportunities in robotic control as on-board compute size poses an architecture search constraint.

Differentiable Architecture Search (DARTS)~\cite{liu2018darts} is a machine learning technique used to automate the process of finding optimal neural network architectures for tasks by introducing a continuous relaxation of the discrete architecture space, allowing gradient-based optimization methods to be used. In~\cite{pmlr-v188-miao22a} DARTS was used for reinforcement learning policies. In~\cite{akinola2021visionary} a  differentiable approach was used for architecture search for robotic learning - the first to deploy a NAS-based neural controller on a robot.  
Efficient Neural Architecture Search (ENAS)~\cite{zoph2016neural} optimizes the architecture search process by sharing parameters across child models, reducing the computational overhead of evaluating multiple architectures. \cite{song2019reinforcement} and \cite{10.1007/978-3-030-80126-7_42} utilize ENAS to find the best-performing architecture for RL tasks.

Another family of methods in NAS is one-Shot Model Architecture Search through Hypernetworks (SMASH)~\cite{brock2018smash}. A primary network (hypernetwork) is trained to estimate the optimal weights for a variable architecture secondary network. Once this hypernetwork is trained, the optimal weights for all architectures in a search space can be estimated, and the one with the best objective can be chosen. The idea of 
 Graph Hypernetworks (GHN) was introduced in~\cite{ deb60d4fcace4f1cb817e737495bfd97}. The computational graph of an architecture is provided as input, and common message-passing techniques akin to those found in GNNs are used to generate the weights of that architecture as its output. GHN benchmarking against other DARTS and ENAS methods shows that it only uses a fraction of the search cost associated with other NAS methods. Following this~\cite{knyazev2021parameter} introduces GHN2, which employs a gated graph network for better generalization of the hypernetwork.

\cite{hegde2023efficiently} introduced Graph Hyper Policies (GHP) that utilized a GHN to estimate the weights of robotic policies for manipulation and locomotion. This was done using off-policy reinforcement learning, specifically, Soft Actor critic \cite{haarnoja2018soft} for locomotion and Hindsight Experience Replay\cite{andrychowicz2017hindsight} with Deep deterministic policy gradients \cite{lillicrap2016continuous} for manipulation. 
For a given architecture graph representation of a network $g$, this network, $h_\theta$, can estimate the policy $\pi_\phi = {h_\theta}(g)$, where the estimated weights are $\phi$.
It was also shown in \cite{hegde2023efficiently} that directly estimated weights of smaller policies were more performant than policies of same same architecture obtained by behavior cloning based distillation methods.  
Since these methods are off-policy, they are extremely sample efficient and can learn to estimate weights for multiple policies with the same number of samples as it would be to learn for a single architecture. 
A drawback for this method though is that it is not time efficient. As noted in the paper, this method had a $\sim$ 5x training time increase. 
This can amount to a large amount of time considering that off-policy methods are already time inefficient as compared to on-policy methods.
Further, this method does not scale well with more compute resources as data collection is not a bottleneck for Q learning.
From a constraint architecture search point of view, searching for architectures for robotic control, hypernetwork-based methods are an alluring option as having multiple options during deployment would reduce experimentation time drastically.

\subsection{Deep Reinforcement Learning for Quadrotor Control}
\label{sec:related_works_drl_quadrotor}
There is significant recent work in the control of quadrotors with direct rotor thrusts by using deep reinforcement learning (DRL).
\cite{hwangbo2017control} investigates stabilizing a quadrotor with hash initialization, and a neural network policy with two hidden layers with 64 neurons in each layer. \cite{molchanov2019sim} can train control policies with minimal prior knowledge about a quadrotor's dynamics parameters and can transfer a single control policy to multiple quadrotor platforms with two hidden layers with 64 neurons in each layer. \cite{lambert2019low} uses model-based DRL for the hover control of a quadrotor (up to 6 seconds with 3 minutes of training data with 2 hidden layers with 250 neurons in each layer). \cite{song2021autonomous} proposes control policies that can achieve 60 km/h on a physical quadrotor by using 2 hidden layers with 128 neurons in each layer. \cite{batra2022decentralized} uses DRL to design decentralized control policies that can fly quadrotor swarms in various scenarios with significant collision avoidance ability in the real world with two encoders, both consisting of 2 hidden layers, with only 16 and 8 neurons, respectively.

For agile tasks, it is desirable for neural network inference to have lower latency than sensing. 
This can become an issue when the sensing modality is complex (such as vision) or goal conditioning needs a larger encoder (such as language).
For agile flight control of a quadrotor, \cite{loquercio2021learning} utilize a RealSense D435i camera for depth sensing, which runs at 30 Hz while their network inference on an onboard NVIDIA Jetson TX2 runs at 25 Hz.

\section{Method}
\label{sec:method}

\subsection{Multi Architecture Proximal Policy Optimization}
\label{sec:hppo}

The method proposed in~\cite{hegde2023efficiently} is off-policy. Such methods tend to be sample efficient, yet time-inefficient in training. To find an on-policy version of~\cite{hegde2023efficiently}, as a first cut, we ran PPO where the policy is replaced with a graph hyper policy estimating policies for randomly sampled architectures, on the halfcheetah environment~\cite{DBLP:journals/corr/BrockmanCPSSTZ16}. This setup is similar to \cite{hegde2023efficiently} but with PPO instead of Soft Actor Critic~\cite{haarnoja2018soft}. 
As the model trained, we evaluated it on a fixed set of architectures. 
We observed that for all architectures, the policies estimated by the graph hyper policy reach the same reward and collapse to a single policy. 
    This is because PPO, being an on-policy algorithm, cannot effectively use data obtained from one architecture to estimate weights for a different architecture.
    This becomes evident on inspecting the equations for PPO from \ref{sec:related_works_ppo}.
    
    Let us denote the entire search space of architectures by $\mathrm{U}$. Let the sampled architectures from this space be $g \sim \mathrm{U}$. 
    In order to use the PPO algorithm for multi-architecture training, we need to substitute $\pi_\theta \leftarrow {h_\theta}(g)$ in these equations, where $h_\theta$ is a graph hypernetwork parameterized by $\theta$, which estimates the weights for architecture $g$. Doing so results in the following equations:

    \begin{center}
        $\begin{aligned}
        r_t(\theta, g)&= \frac{{h_\theta}\left(a_t \mid s_t, g\right)} {{h_\theta}_k\left(a_t \mid s_t, g\right)} \\
        \hat{A}_t^{{h_\theta}(g)}&=\delta_t+(\gamma \lambda) \delta_{t+1}+\ldots+(\gamma \lambda)^{T-t+1} \delta_{T-1} \\
        \text { where } \delta_t&=r_t+\gamma V^{{h_\theta}}\left(s_{t+1}, g\right)-V^{{h_\theta}}\left(s_t, g\right) 
        \end{aligned}$
        \vspace{4pt}\\
        $ \mathcal{L}_{\theta_k}(\theta) = \underset{\substack{g \sim \mathrm{U}\\ \tau \sim {h_\theta}(g)}   }{\mathrm{E}}\left[\min \left(
            \begin{aligned}
                &r_t(\theta, g) \hat{A}_t^{{h_\theta}_k(g)}, \\
                &\operatorname{clip}\left(r_t(\theta, g), 1\pm \epsilon\right) \hat{A}_t^{{h_\theta}_k(g)}
            \end{aligned}            \right)\right]$
    \end{center}

    We see that the importance sampling ratio, advantage estimate, and the value function, are all now conditioned on the current policy's architecture. 
    Since the architecture remains $g$ while estimating all the above values, no mixing of data between architectures must happen.
    
\subsection{Intuition}
\label{sec:theory}
\begin{figure}
        \vspace{2mm}
        \centering
        \includegraphics[width=0.9\columnwidth]{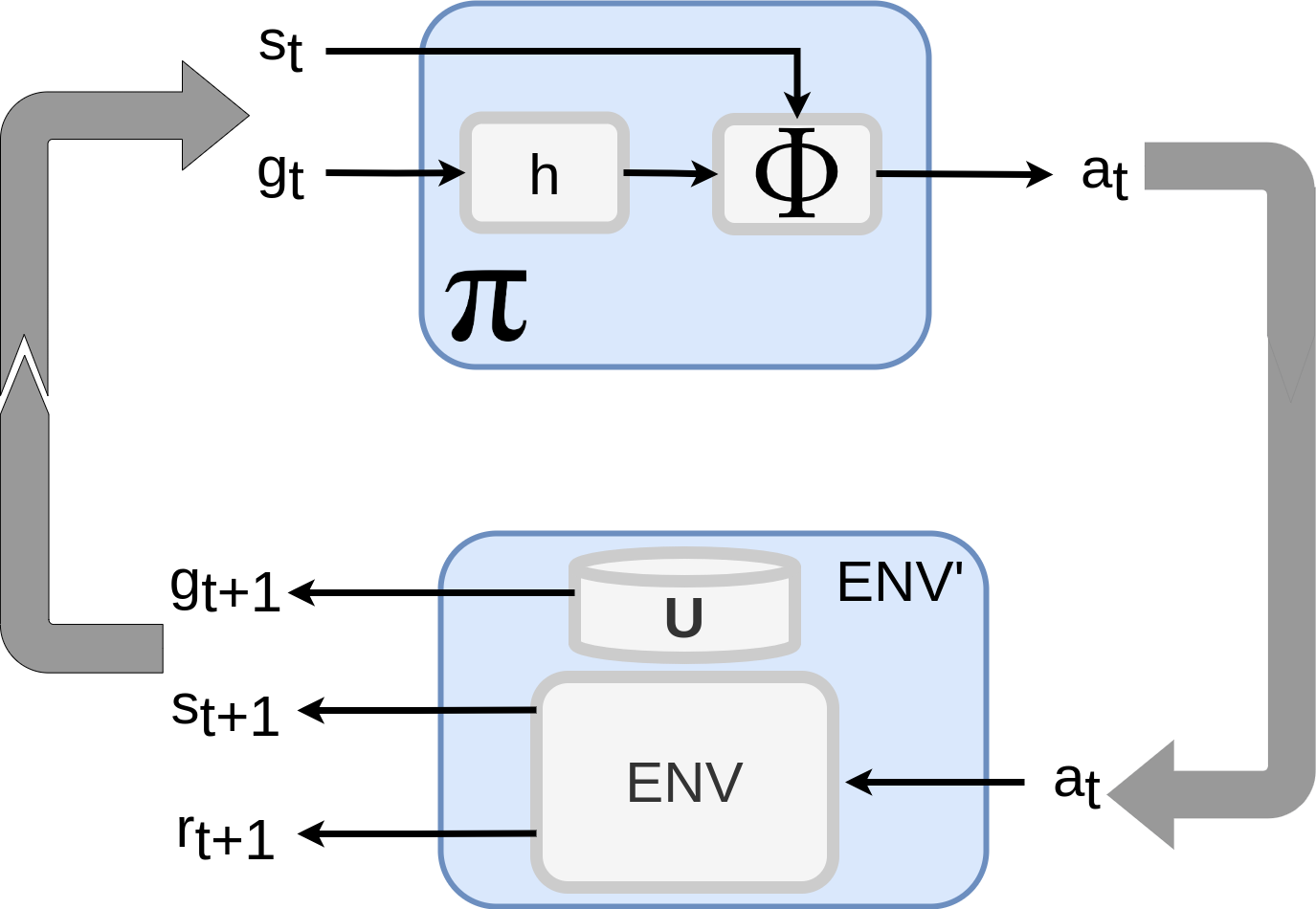}
        \caption{Architecture and State concatenated Markov Decision Process. By augmenting the architecture into the MDP state space, we can train policy RL agents with varying architecture.}
        \label{fig:mdp}
        \vspace{-4mm}
    \end{figure}

    Another way of visualizing the above formulation is by restructuring the underlying Markov Decision Process. 
    We concatenate the randomly sampled architecture graph into the state variable. 
    As shown in figure \ref{fig:mdp}, this allows us to reformulate the policy as the actions sampled from the policy estimated by the hypernetwork for that given combination of graph and state variables. 
    The concatenation of the state and architecture can be seen while estimating the GAE $\hat{A}_t^{{h_\theta}(g)}$, specifically while estimating the state value function $V^{{h_\theta}}\left(s_t, g\right)$. 
    Practically, we condition the critic network of PPO with state and architecture and make sure we use the same architecture's data for the Bellman update.

\subsection{Algorithm}
\label{sec:also}
    Based on these changes we propose HyperPPO. 
    As shown in Algorithm~\ref{algo}, for a given task we start with a predefined architecture space $\mathrm{U}$.
    For every iteration of the algorithm, we sample architecture $g_i$ from the search space.
    For this work, we restrict the search to the architecture space of Multi Layer Perceptrons (MLPs).
    Our architecture search space $\mathrm{U}$ consists of all possible MLPs with four or fewer layers. that can be constructed with the number of neurons in each layer being \{4, 8, 16, 32, 64, 128, 256\}. This gives us 2800 unique architectures.
    We use the same graph hyper policy model as in~\cite{hegde2023efficiently} and estimate policy ${\pi_\phi}_i$ for that architecture.
    We then collect data $\{\mathcal{D}_k\}_i$ using this policy.
    Using this data we estimate GAE ${\hat{A}^{{h_{\theta_k}}(g_i)}}_t$ and the ratio function $r_t(\theta, g_i)$.
    This process can be parallelized for a meta batch size of architectures for faster computation.
    Using these estimates, we then use SGD to optimize the objective $\mathcal{L}_{\theta_k}$ over the hypernetwork weights $\theta$. 
    
    Just like regular PPO for continuous action spaces, actions are sampled from a Gaussian distribution. 
    The mean of the distribution is obtained using the policies estimated by the graph hyper network.
    For standard deviations, we propose two approaches, which lead to two versions of our method.
    HyperPPO-VSD (Vectorized Standard Deviations) constructs a vector of standard deviation arrays, one for each architecture. 
    This enables independent exploration for all architectures. 
    HyperPPO-CSD (Common Standard Deviation) uses a common standard deviation array for all architectures. This reduces computation and converges faster.

    For our method, we utilize vectorized environments. These environments enable parallelization and allow us to sample data for different architectures simultaneously. 
    The larger the number of environments we can run in parallel the better our estimates should be for our objective functions.

    \begin{algorithm}[]
	\caption{HyperPPO} 
	\begin{algorithmic}[1]
            \State input: Initial Hypernetwork parameters $\theta_0$.
            \State input: Clipping threshold $\epsilon$.
            \State input: Architecture Search space $\mathrm{U}$, Meta-batch size $M$.
		\For {$k=1,2,\ldots$}
                \For {$i=1,2,\ldots M$}
                    \State Sample architecture $g_i \sim \mathrm{U}$
                    \State Estimate Policies ${\pi_\phi}_i \leftarrow {h_{\theta_k}(g_i)}$
                    \State Collect trajectories $\{\mathcal{D}_k\}_i$ using policy ${\pi_\phi}_i $
    			\State Estimate GAE ${\hat{A}^{{h_{\theta_k}}(g_i)}}_t$
                    \State Estimate importance sampling ratio $r_t(\theta, g_i)$
                \EndFor
                \State Compute policy update
                    \State \begin{center}
                        \vspace{-5mm}
                        $\theta_{k+1} = arg max_\theta \mathcal{L}_{\theta_k}(\theta)$
                    \end{center}
                \State by taking $K$ steps of minibatch SGD (via Adam)
		\EndFor
	\end{algorithmic} 
    \label{algo}
    \end{algorithm}
    \vspace{-3mm}

\section{Results and Discussion}
\label{sec:expts}

\newcommand\plotimagewidth{86mm}
\begin{figure*}[h]
\centering
\vspace{2mm}
\setlength{\tabcolsep}{2mm}
\begin{tabular}{c c}
\multicolumn{1}{c}{
\includegraphics[width=\plotimagewidth]{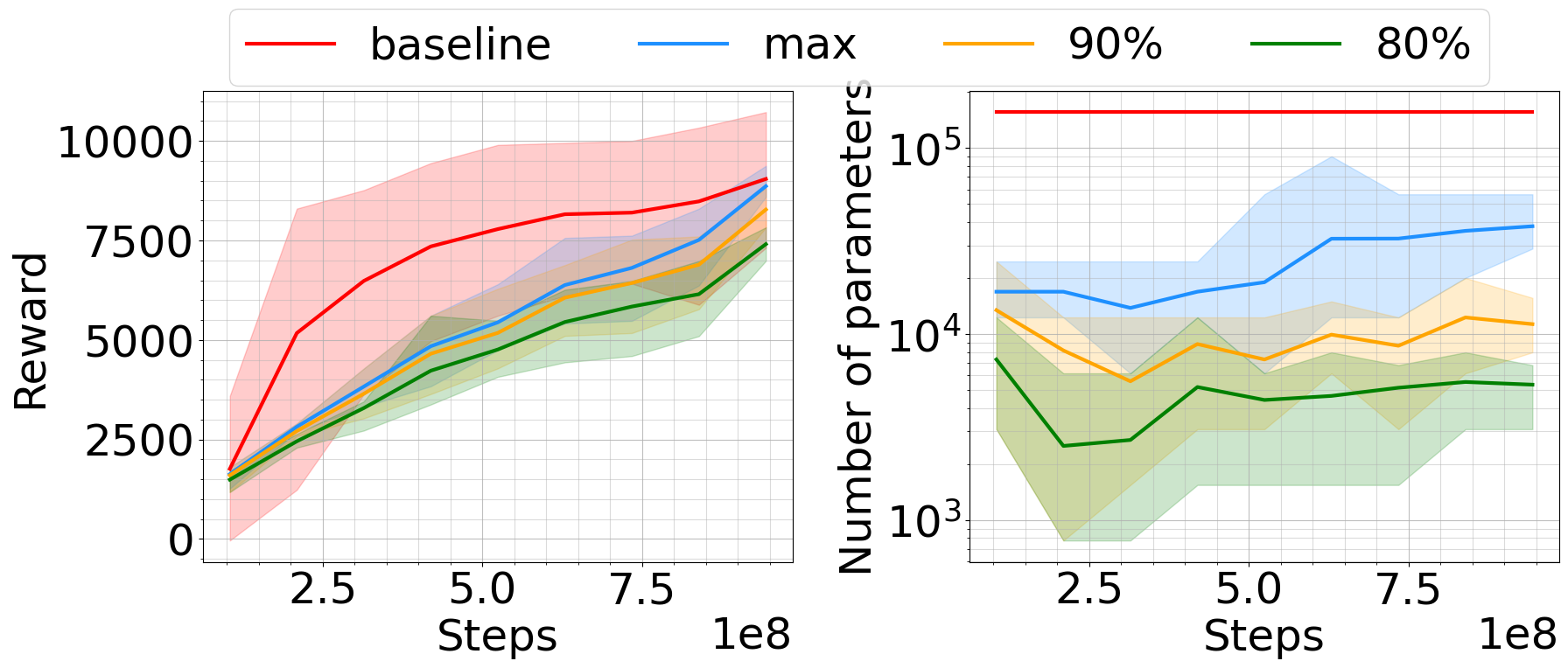}} &   \multicolumn{1}{c}{\includegraphics[width=\plotimagewidth]{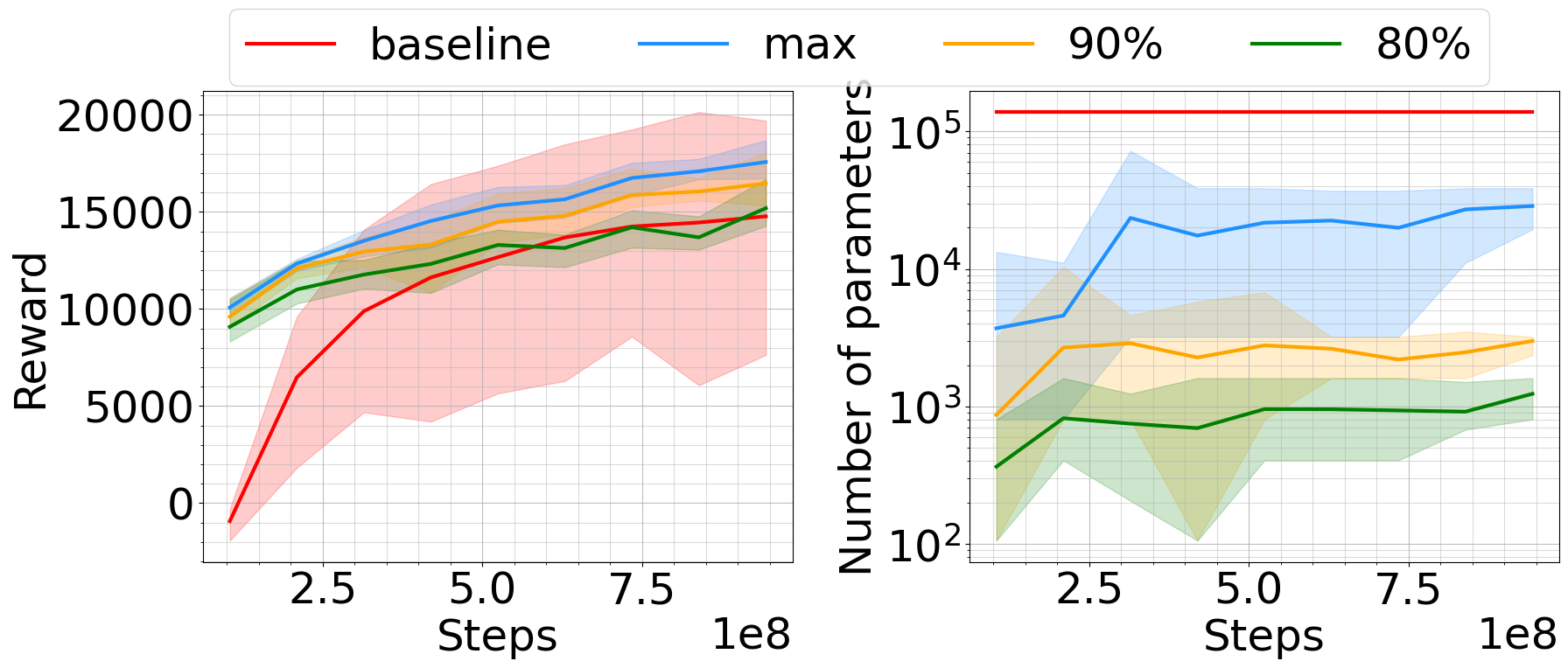}} \\[6pt]

\multicolumn{1}{c}{(a) Ant} &   \multicolumn{1}{c}{(b) HalfCheetah} \\[6pt]

\multicolumn{1}{c}{\includegraphics[width=\plotimagewidth]{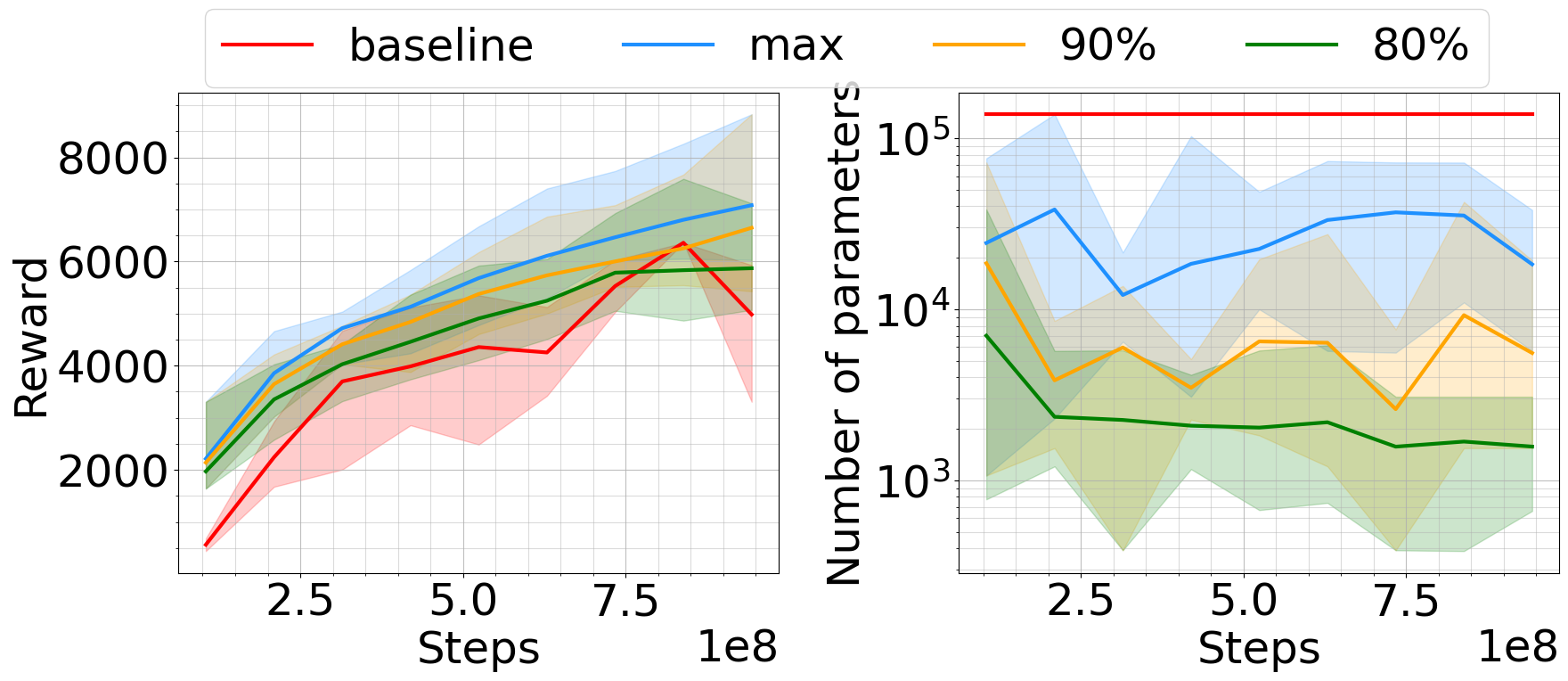}} & \multicolumn{1}{c}{\includegraphics[width=\plotimagewidth]{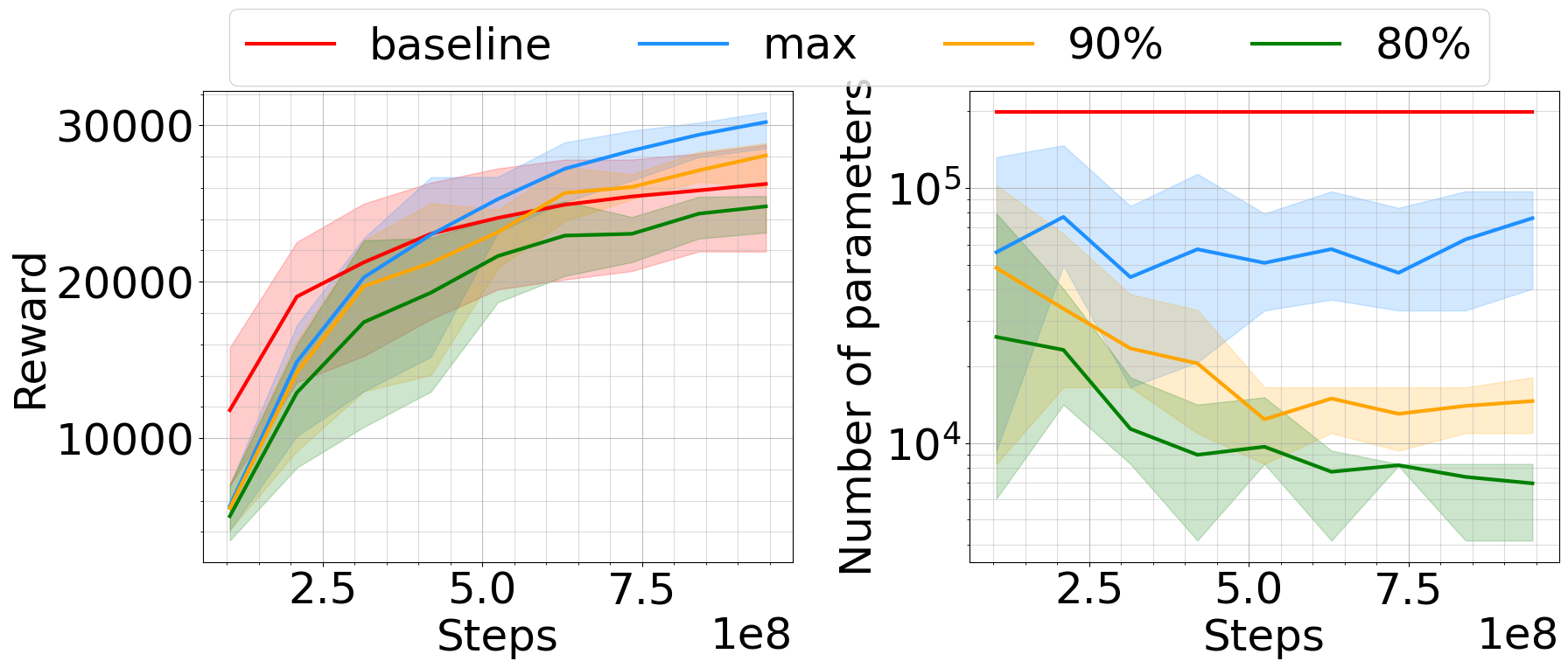}} \\[6pt]

\multicolumn{1}{c}{(c) Walker2D} &   \multicolumn{1}{c}{(d) Humanoid} \\[6pt]

\end{tabular}





\caption{\textbf{Learning smaller networks.} All architectures are evaluated as training progresses. For each pair, \textbf{left}: (max performance, 90\% of max performance, 80\% of max performance, baseline performance) vs training samples collected; \textbf{right}: the minimum number of parameters needed to achieve these levels of performance vs training samples collected.}
\label{fig:percent_plots}
\vspace{-2mm}
\end{figure*}

To implement our method, we use the Sample Factory~\cite{petrenko2020sf} package. Its efficient design enables us to parallelize data collection and train Graph Hyper Policies quickly. 
The experiments are carried out on standard locomotion tasks that have been implemented on Brax~\cite{freeman2021brax} and Mujoco~\cite{todorov2012mujoco}. We also train on the quadrotor simulator described in QuadSwarm~\cite{huang2023quadswarm}.
All experiments were run 4 seeds at a time on an AWS g4dn.12xlarge instance with 48vCPU, 4 Telsa T4 GPUs and 192 GB RAM. 

\subsection{Ablations}
\label{sec:ablations}



\begin{figure}
    \centering
    \includegraphics[width=\columnwidth]{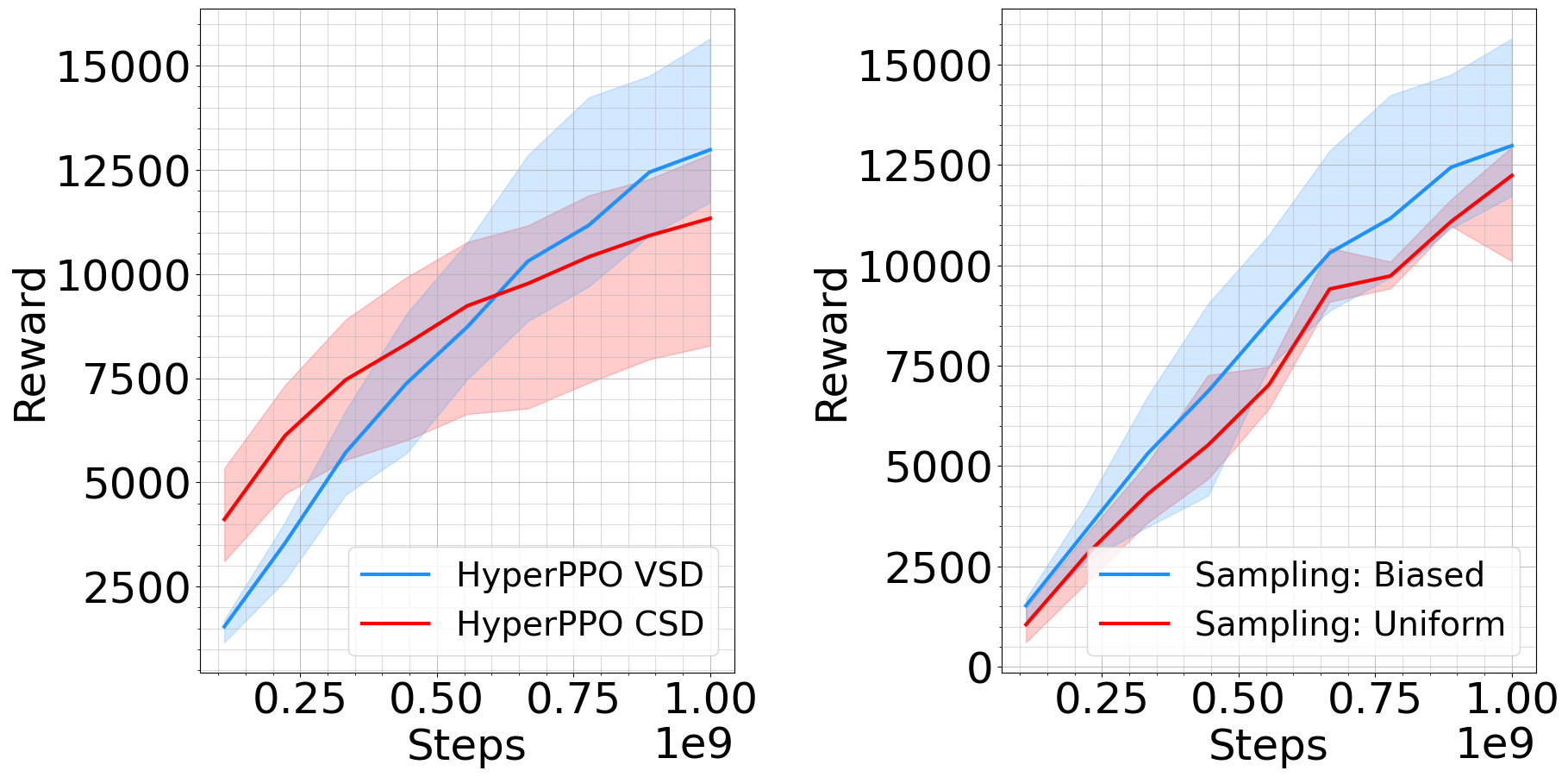}
    \caption{\textbf{Ablations.} Average reward across all architectures during training. \textbf{Left:} Action Standard Deviation; \textbf{Right}: Architecture Sampling.  }
    \label{fig:ablation_std}
        \vspace{-5mm}
    \label{fig:ablation}
\end{figure}

For our ablations, we train on the Humanoid task in Brax for 1 billion steps for 8 seeds. 
We simulated 4096 environment instances in parallel and ran for approximately 200 minutes.
Every few steps, we evaluate the performance of policies estimated by the GHP for every architecture in the search space. 
To estimate the quality of all architectures we find the average reward across all architectures.

\subsubsection{Vectorized Standard deviations}
\label{sec:ablations_vec}
First, we analyze the performance of HyperPPO with VSD and CSD. 
Figure~\ref{fig:ablation} shows this for both CSD and VSD.
We see that with CSD, the average reward grows faster initially. This is because the standard deviation converges faster with CSD. But with more training, we see that VSD eventually achieves a larger reward. As mentioned in~\ref{sec:ablations_vec}, we believe this is because individual exploration for each architecture can eventually obtain better performance.
Therefore we suggest using the VSD when massively parallel environments such as Brax or IsaacGym~\cite{makoviychuk2021isaac} are available.

\subsubsection{Architecture Sampling}
\label{sec:ablations_arch_samp}
During experimentation, we first implemented the uniform architecture sampling as described in~\cite{hegde2023efficiently}. On further analysis, we found that the graph hyper policy has a learning bias toward deeper network architectures. We believe this is because there are fewer shallower architectures than deeper ones. To compensate for this effect, we sample architectures with their sampling probability inversely proportional to the number of layers. We shall call this biased sampling.

We run HyperPPO-VSD with both modes of architecture sampling. From figure~\ref{fig:ablation}, we can see that with biased sampling, we obtain better performance. Further, smaller networks gained a bigger performance boost with biased sampling, since more of these were considered during training. Therefore, for all other experiments in this paper, we set the architecture sampling mode to biased sampling.


\subsection{Scaling HyperPPO}
\label{sec:time_eff}
\begin{figure}
    \centering
    \includegraphics[width=\columnwidth]{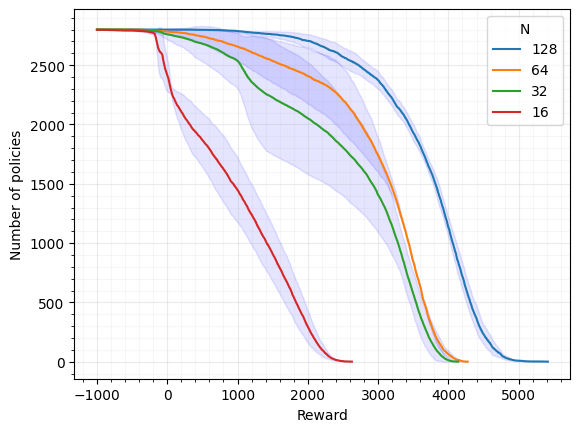}
    \caption{\textbf{Scaling HyperPPO}: With more environment instances, the performance of all architectures increases. N represents the number of parallel environment instances.}
    \label{fig:compare}
    \vspace{-5mm}
\end{figure}

Here, we show that HyperPPO can scale up to provide better results with more computation. 
We train HyperPPO-CSD on the Mujoco halfcheetah task for 5 hours while varying the number of environment instances from which data are sampled. We run this experiment over 5 seeds, and at the end of the experiment, we evaluate every architecture in the search space.
Figure~\ref{fig:compare} shows us the distribution of performance over all unique architecture policies estimated by GHP. This plot is similar to those used to evaluate policy data sets in \cite{batra2023proximal,hegde2023generating}. The x-axis is the policy's accumulated reward, while the y-axis represents the number of policies with reward greater than x. N represents the number of environment instances from which data are sampled. We can see that scaling up the algorithm with more parallel environments in HyperPPO with more computation can provide a better collection of policies over the same time. 

\subsection{Brax benchmarks}
\label{sec:brax_bench}

Having shown that our method scales with performance, we benchmark HyperPPO-VSD on GPU-accelerated Brax environments.
We use 4 locomotion tasks, namely, humanoid, ant, halfcheetah, and walker2d. On each task, we train for 1 billion state transition steps and show results across 8 seeds. 
During training, every few steps, we evaluate the GHP on every architecture in the search space. From this evaluation, we identify architectures that provided the highest reward, the smallest architectures that provided 90\% of the highest reward, and the smallest architectures that provided 80\% of the highest reward.
We call these max, 90\%, and 80\% architectures respectively. As a baseline, we train regular PPO also implemented on Sample Factory with the same hyperparameters, with 3 hidden layers with 256 neurons each. This is a common choice of model architecture for these locomotion tasks.
Figure~\ref{fig:percent_plots} shows the results of this experiment. For each task, the left plot depicts rewards attained by the max, 90\%, 80\% architectures, and the baseline.  The right plot shows the size of these architectures on a log scale. 
For all tasks, we see that the number of parameters required to achieve 90\% and 80\% of maximum performance reduces considerably.

Further, by taking the average reward over all seeds, we identify 80\% architectures for each task as (64) for halfcheetah, (64) for walker2d, (32) for humanoid, and (64) for Ant. These are all single hidden layer architectures with either 64 or 32 neurons in them. We trained policies with these architectures with regular PPO and compared their performance with policies of the same architectures estimated by the GHP in HyperPPO-VSD.
Table~\ref{tab:mytable} shows that the policies estimated by the GHP obtain considerably more reward on the Halfcheetah, Walker2d, and Humanoid tasks, while the performance is comparable on the Ant task, figure \ref{fig:percent_plots} suggests that the model has not yet converged for Ant.


\begin{table}[h]
    \vspace{5mm}

    \centering
    \begin{tabular}{c|c|c|c}
        \multicolumn{1}{c}{Task} & \multicolumn{1}{|c}{\shortstack{80\%\\Architecture}} & \multicolumn{1}{|c}{PPO (x$10^2$)} & \multicolumn{1}{|c}{HyperPPO (x$10^2$)} \\
        \hline
        Halfcheetah & [64] & $80.30 \pm 49.23$ & $\mathbf{144.80 \pm 13.36}$ \\
        \hline
        Walker2D & [64] & $19.84 \pm 7.18$ & $\mathbf{58.50 \pm 6.64}$ \\
        \hline
        Humanoid & [32] & $182.85 \pm 25.35 $ & $\mathbf{207.69 \pm 49.12}$ \\
        \hline
        Ant & [64] & $71.88 \pm 11.46$ & $70.49 \pm 8.85$ \\
    \end{tabular}
    \caption{Comparison of small policies}
    \label{tab:mytable}
    \vspace{-4mm}
\end{table}

These results show that HyperPPO-VSD can provide multiple architecture policies with the same sample complexity as a single PPO run, and further provides higher performing smaller policies than its regular PPO counterparts.
We believe this increase in performance has two reasons: (a) Better exploration: The policies are now more stochastic with HyperPPO-VSD probabilistically choosing different action distributions during data collection. (b)  Distillation between architectures: Gradients to the hypernetwork from data of larger architectures can improve policies estimated for smaller architectures.

\subsection{Quadrotor Drones}
\label{sec:quad_bench}
\begin{figure}
\vspace{2mm}
    \centering
    \includegraphics[width=\columnwidth]{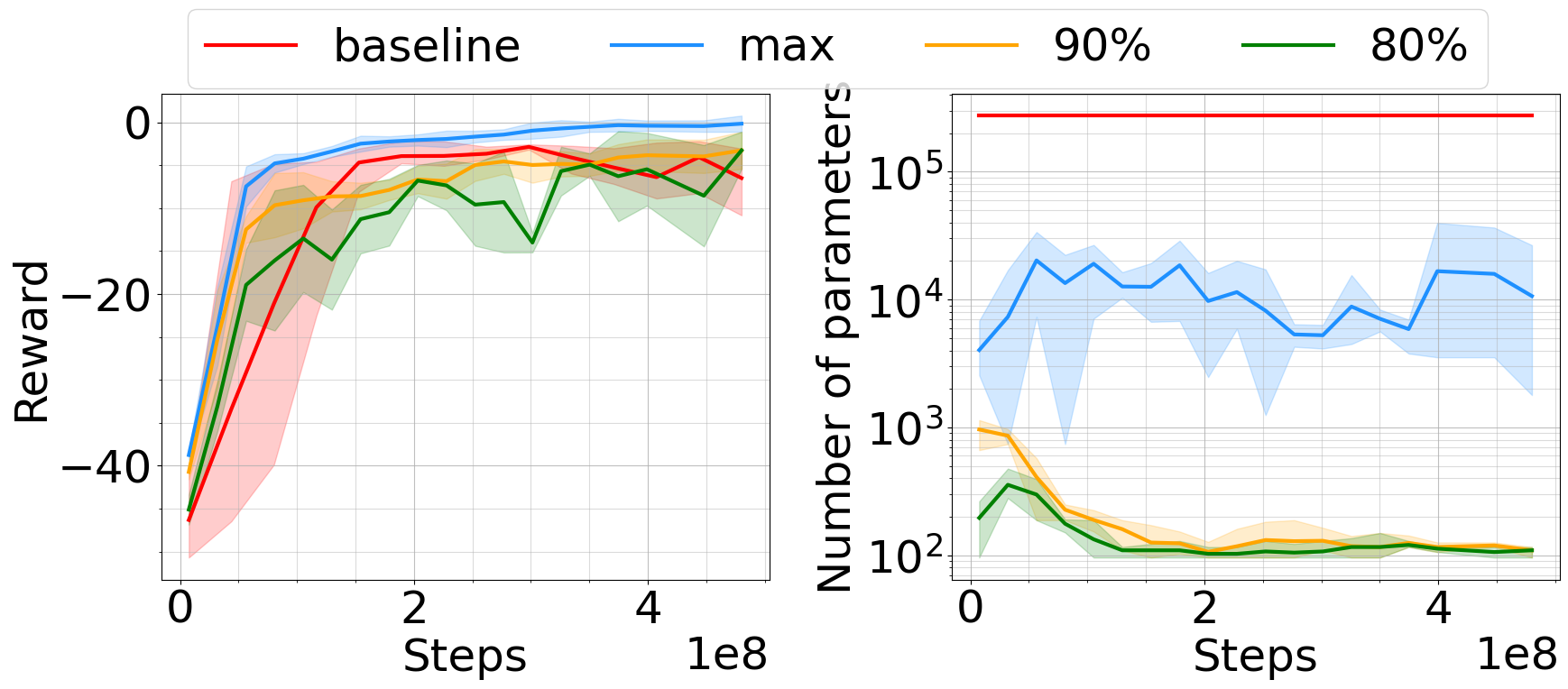}
    \caption{Analysis of Quadrotor Training}
    \label{fig:drone_results}
\end{figure}

\begin{figure}
    \centering
    \includegraphics[width=\columnwidth]{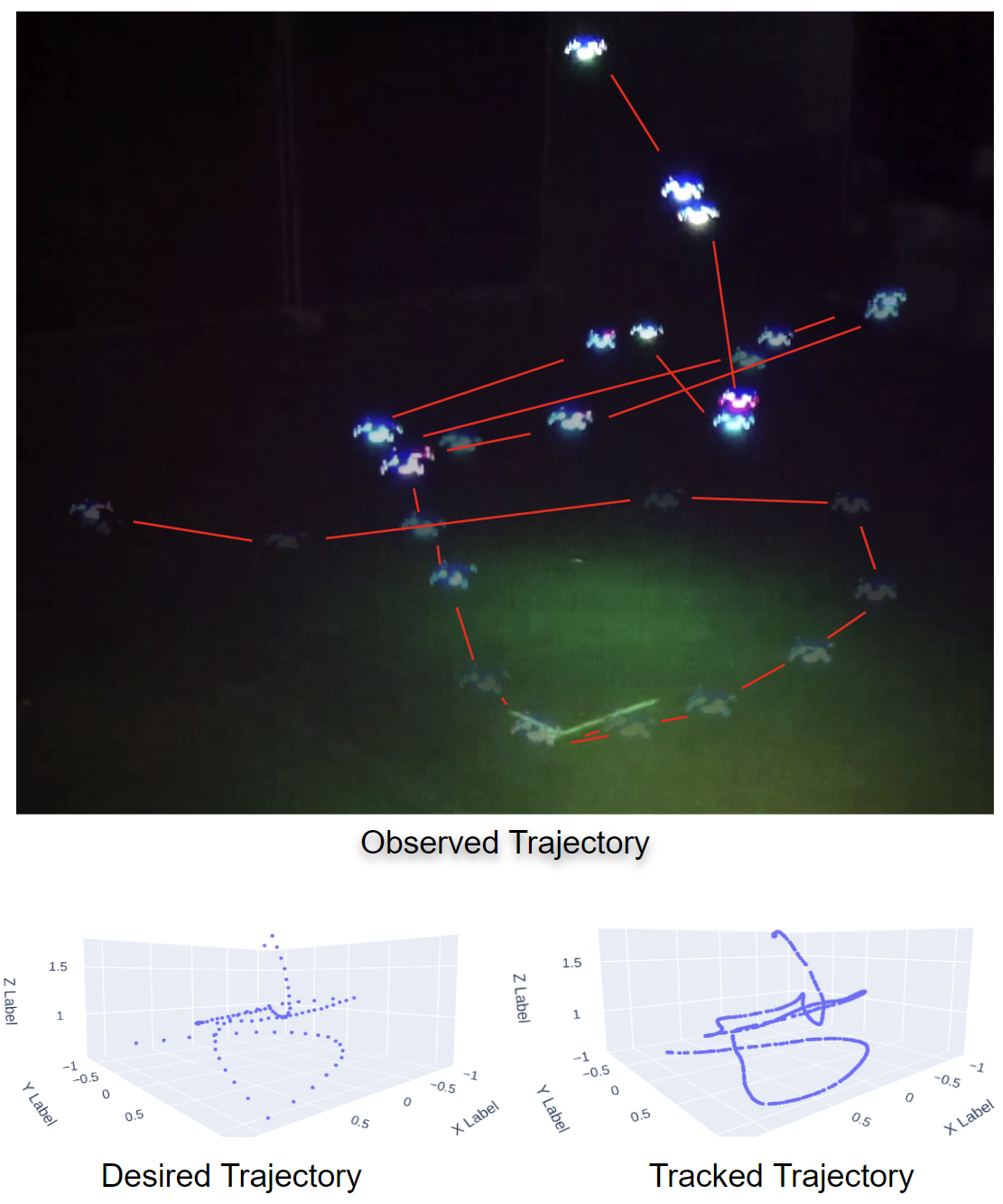}
    \caption{Evaluating a single layer 4 neuron network estimated by HyperPPO-CSD on the Crazyflie2.1. \textbf{Left:} The desired trajectory created with a random bezier curve. \textbf{Right:} Actual trajectory of the drone. \textbf{Top:} Frames stacks of the actual footage of drone flight.}
    \label{fig:drone_real}
    \vspace{-4.5mm}
\end{figure}

We train HyperPPO-CSD on the Quadrotor environment designed for a Crazyflie 2.1, QuadSwarm~\cite{huang2023quadswarm}. The Crazyflie 2.1 is a severely compute-constrained quadrotor with an onboard microcontroller running at 168MHz with 168 Kb RAM.  
We train the control policy in simulation on a mixture of single drone goal-based scenarios~\cite{batra2022decentralized} (static goal, dynamic goal, random 3D Lissajous trajectory tracking, and random 3D Bezier curve trajectory tracking), for 500 million state transition steps, and we zero-shot transfer our control policy to the physical Crazyflie quadrotor. 
We test our control policies on the Bezier curve trajectory tracking on the physical Crazyflie quadrotor, one of the most challenging scenarios in the simulation, to showcase the flying performance of our control policy. 
As a baseline, we train a policy with architecture (512,512) (i.e., two hidden layers with 512 neurons each), with the same hyperparameters and scenarios. 
Similar to Figure~\ref{fig:percent_plots}, we analyze the training performance in Figure~\ref{fig:drone_results}. 
We see that the best-performing architecture estimated with HyperPPO-CSD achieves more reward than the baseline, whose performance is comparable to that of 80\% architectures.
Across seeds, for this task, we identified the 80\% architecture as (4) (i.e., a single hidden layer 4 neuron network). This small policy was estimated at the end of training and deployed on the Crazyflie.
For evaluating the physical deployment performance, we generate a random 3D Bezier curve as the desired trajectory and use the neural network to control rotor thrusts, to track this trajectory. From Figure~\ref{fig:drone_real} we see that the quadrotor is capable of tracking the desired trajectory with a HyperPPO estimated neural network, with high success rates.
If we wanted to test a different architecture for physical deployment, instead of retraining a new network from scratch, we can estimate the weights for that architecture with one inference step of the trained GHP model. 

While we maintain sample efficiency, we note that a limitation of our method is a $\sim$ 2-3x training time increase as compared to regular PPO.
At present, we limit ourselves to Multi-Layer Perceptrons, however, we plan to experiment with architecture search spaces with different types of networks such as CNNs, LSTM, and Transformers in the future.
Finally, identifying the performance of a candidate architecture involves estimating it with the GHP and evaluating it with a rollout. Identifying the desired architecture algorithmically during training is a possible future avenue for this work.

\section{Conclusion and Future Work}
\label{sec:con}

We present HyperPPO, an on-policy algorithm that learns multiple architecture policies simultaneously. 
We show that the algorithm is fast, sample efficient, and scales with added computation.
We provide two versions: HyperPPO-VSD, which can be used when data collection is accelerated; and HyperPPO-CSD, which can be used when computation is limited and for faster convergence.
We show that on Brax benchmarks, HyperPPO-VSD can quickly estimate thousands of working policy architectures, and the estimated small policies outperform PPO on most tasks.
Finally, we show that small policies estimated by HyperPPO-CSD can be successfully deployed on an actual compute-constrained platform - the Crazyflie - for neural control.




\newpage
\bibliographystyle{IEEEtran}
\bibliography{refs}

\end{document}